\documentclass[12pt]{elsarticle}
\usepackage{multirow}

\usepackage{amssymb}
\usepackage{amsthm}

\usepackage[left=2.5cm, right=2.5cm, top=4cm, bottom=3cm, footskip=0.5cm]{geometry}

\journal{ }

\begin{document}

\begin{frontmatter}

\title{Inter-Sense: An Investigation of Sensory Blending in Fiction}
\date{ }

\author[1,2]{Roxana Girju}
\author[2,*]{Charlotte Lambert}

\address[1]{Department of Linguistics and Beckman Institute, University of Illinois at Urbana-Champaign, Urbana IL 61802}
\address[2]{Department of Computer Science, University of Illinois at Urbana-Champaign, Urbana IL 61802}
\address[*]{Corresponding author: girju@illinois.edu}

\begin{abstract}
This study reports on the semantic organization of English sensory descriptors of the five basic senses of sight, hearing, touch, taste, and smell in a large corpus of over 8,000 fiction books. We introduce a large-scale text data-driven approach based on distributional-semantic word embeddings to identify and extract these descriptors as well as analyze their mixing interconnections in the resulting conceptual and sensory space. The findings are relevant for research on concept acquisition and representation, as well as for applications that can benefit from a better understanding of perceptual spaces of sensory experiences, in fiction, in particular, and in language in general.
\end{abstract}


\end{frontmatter}


\section{Introduction}

Language captures to different degrees people’s sensory perceptions. In spite of significant progress in language and perception in the last decades, little is known about the descriptive coverage of the basic five senses (sight, hearing, touch, taste, and smell) and their inter-relations \cite{Winter2019}, in spoken as well as in written language.
We argue that a good starting point in this direction is fiction. This is because emotions are grounded in sensations and fiction writers weave their characters’ feelings by describing their sensory experiences in intricate ways and with different degrees of metaphor activation. Readers thus can experience emotions through sensory blends, especially if writers use language that activates more than one sense at a time. The most common expression of sensory empathy is based on visual imagery - people like to read because reading allows them to enter the character’s mind and imagine life from that person’s perspective \cite{Zunshine2006}. Readers expect to see, feel, or smell something familiar that helps them experience the emotion or the sensory event the author is describing. Stories often feature such sensory details that trigger the reader's own response.

One important yet largely unexplored line of inquiry in the field of emotion and empathy is how emotions and sensory mixing can effectively build empathy. Although empathy does not necessarily require sensory stimulation, it predicates on peoples' perceptions, their cognitive states, as well as social context. Thus, empathy is fundamentally a perceptual process. As fiction writer Flannery O’Connor writes in \emph{Mystery and Manners}, "[..] the nature of fiction is in large measure determined by the nature of our perceptive apparatus. The beginning of human knowledge is through the senses, and the fiction writer begins where human perception begins. He appeals through the senses, and you cannot appeal to the senses with abstractions" \cite{OConnor1969}. There is a multitude of ways mental images and interoceptive processing can help in integrating different kids of information and expand our way of knowing the world around us. Such integration is in reality a complex process partly because "most imagery is, de facto, not just multidimensional but also multisensory" \cite{Starr2013}.
Successful writers take advantage of this perceptual apparatus.

Readers do not respond to a story exactly the same way and the experience of thinking and feeling varies from person to person \cite{Otis2015}. 
Yet, we believe that empirical studies of such sensory descriptors and their contexts can shed light into the various strategies writers use to best immerse the reader into the characters’ worlds. Such studies will provide a better understanding of the process of combining sensations and the best ways language can engage or move the readers so that we better relate our experiences to those of others. To the best of our knowledge, to date, there is no systematic analysis of writers’ narrative strategies prompting readers to blend the senses for enhanced reader empathy. Although there have been notable recent attempts in computational linguistics, digital humanities, and psychology to automatically extract textual descriptors of sensory experiences  \cite{Brate-etal2020,Iatropoulos-etal2018,Horberg-etal2020}, such research and technology are in their infancy. Most studies have focused on one individual sense, and just a few attempted to combine some of the senses \cite{Scarry1999,Winter2019}. 

We add our own contribution to this growing body of emergent research reporting on a set of findings of a large project that focuses on writers’ narrative strategies to prompt readers to use and combine senses thus creating sensory images that bring characters and scenes to life. 
This paper reports on a component of this research -- namely, the identification and semantic organization of English sensory descriptors in a large corpus of over 8,000 fiction books extracted from Project Gutenberg (www.gutenberg.org/). To investigate a broader range of sensory elements, we employ a large-scale data-driven approach based on distributional-semantic word embeddings \cite{mikolov2013}. Specifically, our aims are: 
\begin{enumerate}
\item to explore to which extent and along what dimensions our distributional approach helps in acquiring the vast and diverse set of contextual descriptors related to the five basic senses (sight, hearing, touch, taste, and smell), and 
\item to analyze their mixing interconnections in the resulting conceptual and sensory space. Specifically, we want to explore which of the five senses tend to occur alone and which are most often combined and how.
\end{enumerate}
Our data visualization results catch a glimpse of the perceptual spaces of sensory experiences in fiction narratives in particular, and in language in general. 
The findings are relevant for research on concept acquisition and representation, as well as for applications that can benefit from a better understanding of perceptual spaces of sensory experiences. For fiction in particular, such line of research may lead to technology that helps inform creative writers on ways to use language empathically to engage readers.

\section{Synesthesia in Literature}
Our five senses – sight, hearing, touch, taste and smell – seem to operate independently, as five distinct modalities of perceiving the world around us. In reality, however, intersensory integration happens constantly enabling the mind to  make sense of its surroundings, even when these senses are not directly perceived.
As Russian novelist Anton Chekhov once said, "Don’t tell me the moon is shining; show me the glint of light on broken glass."\footnote{This passage is attributed to Anton Chekhov as a literary advice he gave to his brother in a letter written in 1886. The English translation of the letter was performed by Avrahm Yarmolinsky \cite{Yarmolinsky1954}.}. When the
reader smells and hears the hot, crispy apple pie freshly baked by the character, sees and touches the smooth, springy table cloth, and hears the purring cat coiled in the corner of the room, the scene becomes real.

The word synesthesia usually refers to a psychological or neurological condition in which sensory stimuli from one sense are mixed up with those of another. For example, some neurological synesthetes can hear color, or smell shapes, or see time in space. The term, however, is also frequently used in literature, referring to a technique of cross-sensory metaphor or ‘intersense analogy’ \cite{OMalley1957} where "perceptions from two different sensory modalities are blended together, for effect" \cite{Duffy2013}. 
Popularized by French symbolists, such as Charles Baudelaire, Arthur Rimbaud and Paul Verlaine, synesthesia is a common literary technique, used as a way to heighten and clarify the symbolic imagery in poems and fiction. Literature abounds in cross-sensory metaphors, and such technique has been explored in numerous studies, including influential works by George Lakoff \cite{Lakoff-Johnson1981}, Lawrence Marks \cite{marks2014}, and Glenn O’Malley \cite{OMalley1957}. However, in literature we don't "take synaesthesia in the strict sense of psychology; that is to say, not with every sound does the poet really see a distinct color; but the impression evocated by the sound or sounds reminds the poet of a similar impression called forth by color. He does not see but thinks color." \footnote{This quote is attributed to Erika Siebold's German study on synesthesia, as cited in Ruddick's "Synesthesia in Emily Dickinson's Poetry" \cite{Ruddick1984}.}.
Probably one of the most popular examples of synesthesia in literature is from Dante Alighieri’s ‘The Divine Comedy’. There, he refers to the place "where the sun is silent"  — meaning, the place the sun cannot be seen -- a lifeless, cold, and colorless place. 

Synesthesia in fiction has been mostly used as a rhetorical device that describes or associates one sense in terms of another (most often as a simile), a form of sensory perception that prompts the reader to go beyond their default understanding of the basic senses.
In this research, however, we use the term more broadly, specifically to refer to ‘sensory blending’ of sense perceptions in text – as an interconnection of senses that tend to occur in similar contexts. In this study, we attempt to address these issues by following the distributional hypothesis \cite{Harris54} which states that the meaning of a word is derived from the linguistic contexts in which it occurs. We use this hypothesis to identify sensory descriptors of different sense combinations that tend to co-occur in fiction writers' books.

Although experienced writers skillfully blend senses for better reader immersion, the mechanism which produces this effect still remains a poorly understood topic. This study aims to narrow the current gaps in our understanding of sensory perception and blending in fiction. We believe that large-scale data-driven approaches like ours can provide us with fresh new insights into how and where the different sensory modalities interact in language, 
and potentially how sensory perception develops in fiction.

\section{Approach}

In this paper, we focus on all five basic senses (sight, hearing, touch, smell, taste) as they group the sensory events at the basic level of categorization. 

We explored and report here our findings on these sensory spaces across the five perception modalities. Below we explain the data collection and processing, the computational modeling of the sensory spaces, and present and evaluate the results.

\subsection{Data}

\begin{figure}
    \centering
    \includegraphics[width=0.8\linewidth]{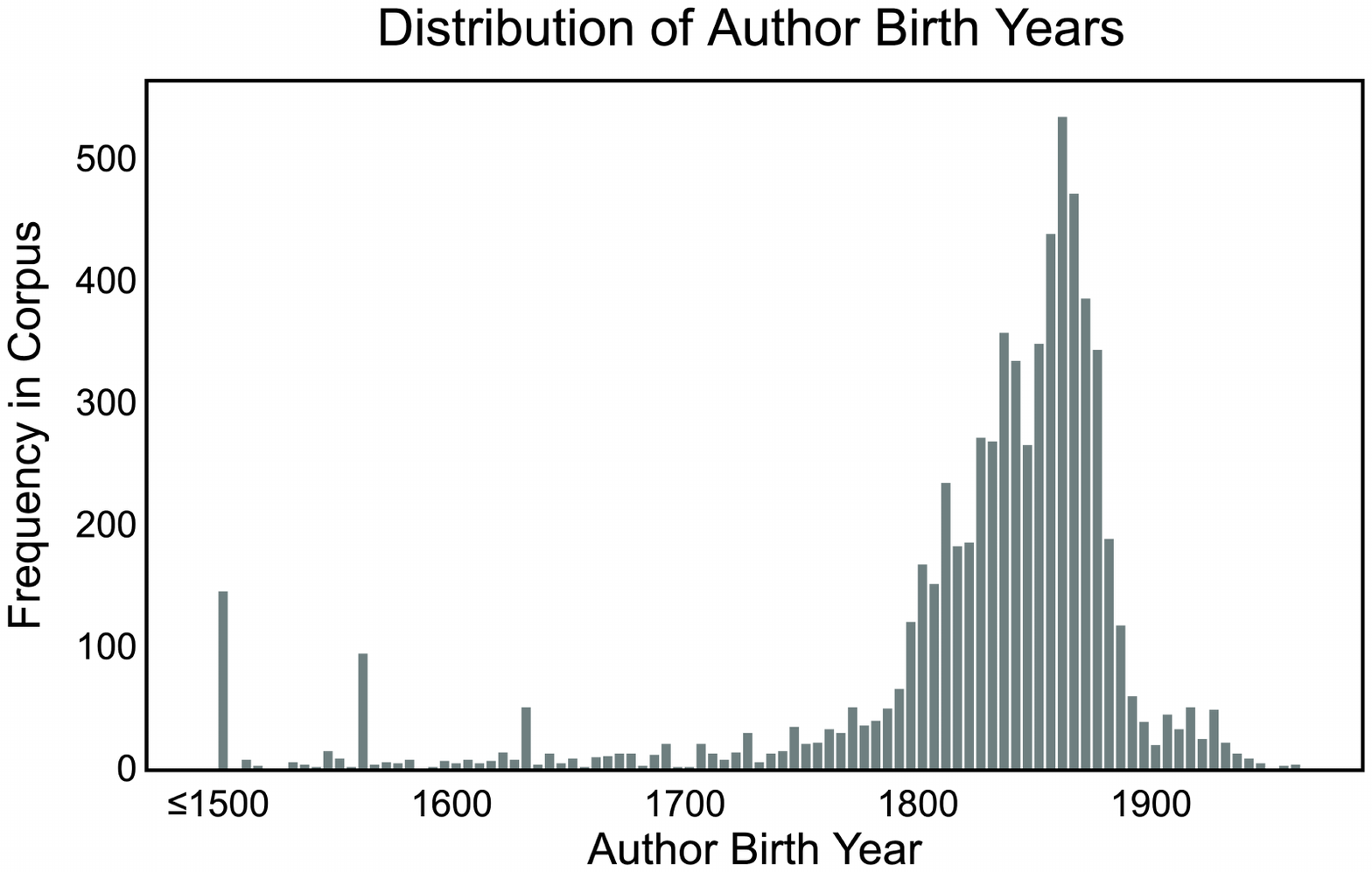}
    \caption{The distribution of author birth years based on the considered fiction corpus.}
    \label{fig:author-birth-years}
\end{figure}

We extracted 8,763 fiction books, categorized by genre and subject. For easy access, we used the Project Gutenberg data (up to June 2016) from the \emph{Gutenberg, dammit} (v0.0.2: 2018-08-11)\footnote{The \emph{Gutenberg, dammit} (https://github.com/aparrish/gutenberg-dammit) was created by Allison Parrish and licensed under the Creative Commons Attribution-ShareAlike 4.0 International License (https://creativecommons.org/licenses/by-sa/4.0/)}. This is a corpus repository of plain text files in Project Gutenberg with consistent metadata associated with each manuscript. We selected all the books in English containing the word "Fiction" in the "Subject" category (i.e., "England – Fiction", "Historical Fiction", "Fiction"). Since some books were listed under multiple genres, we made sure that at least one genre satisfies the fiction requirement. Table~\ref{tab:author-freqs} reports the 20 authors whose written works appear most frequently in the corpus along with the 20 most frequent genres listed. Out of the 8,756 books in the corpus, there are 4,495 unique authors (including those labeled \emph{None Available}).
Additionally, as publication dates were inaccessible in the metadata provided by the \emph{Gutenberg, dammit} repository, we present  data on the distribution of author birth years in our corpus to approximate the time period during which the fiction novels in the corpus were published.  Figure~\ref{fig:author-birth-years} shows the distribution of 6,711 books with available author birth dates. All instances of authors born before 1500 are binned together, however the full range of author birth years for the corpus is 750 BCE to 1961.
All extracted fiction books were tokenized and part of speech tagged using the Penn Treebank style \cite{marcus-etal-1994-penn}.
We also removed stop words, including all morphological variations of light verbs (i.e., \emph{to be, have, go, come}, and \emph{make}) which would not contribute to the semantic sensory space in our distributional approach.

\begin{table}[ht]
\centering
\begin{tabular}{llrlr}
\hline
& \multicolumn{2}{c}{\textbf{Top 20 Authors}} & \multicolumn{2}{c}{\textbf{Top 20 Genres}} \\
{} &                             Author &  Frequency &                                  Genre &  Frequency\\
\hline\hline
\textbf{0}  &                            Various &     433 &                                Fiction &       352 \\
\textbf{1}  &                                 \textit{None Available}   &     257 &                          Short stories &       276 \\
\textbf{2}  &                          Anonymous &     115 &                        Science fiction &       242 \\
\textbf{3}  &                William Shakespeare &      86 &                                 Poetry &       125 \\
\textbf{4}  &                    George Meredith &      50 &                      Adventure stories &       117 \\
\textbf{5}  &                       Samuel Pepys &      44 &          Detective and mystery stories &       109 \\
\textbf{6}  &                   Honoré de Balzac &      37 &                           Love stories &        98 \\
\textbf{7}  &             Robert Louis Stevenson &      35 &                     Historical fiction &        84 \\
\textbf{8}  &                        Georg Ebers &      33 &                                 Essays &        65 \\
\textbf{9}  &               William Dean Howells &      32 &                     England -- Fiction &        65 \\
\multirow{2}{*}{\textbf{10}} &                    \multirow{2}{*}{Charles Dickens} &      \multirow{2}{*}{31} &   English wit and humor --  &        \multirow{2}{*}{64} \\
&&& Periodicals \\
\textbf{11} &                         Mark Twain &      29 &                        Western stories &        60 \\
\multirow{2}{*}{\textbf{12}} &        T. S. (Timothy Shay)  &      \multirow{2}{*}{26} &     Man-woman relationships --  &        \multirow{2}{*}{60} \\
& Arthur & & Fiction & \\
\multirow{2}{*}{\textbf{13}} &  Baron Edward Bulwer  &      \multirow{2}{*}{25} &    \multirow{2}{*}{Conduct of life -- Juvenile fiction} &        \multirow{2}{*}{57} \\
& Lytton Lytton \\
\textbf{14} &                         Bret Harte &      23 &                               Comedies &        55 \\
\textbf{15} &                       Marie Lebert &      22 &                            Fairy tales &        53 \\
\textbf{16} &                   William Le Queux &      22 &      Illustrated periodicals -- France &        51 \\
\textbf{17} &                     Gilbert Parker &      21 &                                  Drama &        48 \\
\textbf{18} &                        Jules Verne &      21 &         Encyclopedias and dictionaries &        44 \\
\multirow{2}{*}{\textbf{19}} &       H. G. (Herbert George)  &      \multirow{2}{*}{20} &  Statesmen -- Great Britain -- &        \multirow{2}{*}{44} \\
& Wells & & Diaries \\
\hline
\end{tabular}
    \caption{Top 20 most frequent authors and genres in the corpus.}
    \label{tab:author-freqs}
\end{table}

In order to focus the research, we compiled five lists of seed words, one non-overlapping list per each of the basic five senses. Specifically, we manually chose between 15 and 25 representative words considered to be commonly used to describe each of the five senses. For this, we relied on WordNet \cite{WordNet}, a freely available general purpose lexical database of semantic relations between words. WordNet links word senses into semantic relations which have been very useful for the purpose of this study. In selecting our seed words, we started with basic concepts identifying the five modalities  (i.e., \emph{see} (sight), \emph{hear} (hearing), \emph{touch} (touch), \emph{taste} (taste), \emph{smell} (smell)) and added words connected to them through semantic relations like hypernymy, hyponymy, and morphologically related words. Each of the resulting lists was then expanded considering all the morphological variations of each word, thus generating about 150 - 250 seed words per sense, all tagged with part of speech information (e.g., \emph{smell}\_v; \emph{smells}\_v; \emph{smelled}\_v; \emph{smelling}\_v; \emph{smell}\_n; \emph{smells}\_n).
 Table~\ref{table-seeds} shows a sample of ten of the original seeds for each sense modality.
 
 We are fully aware that such a general, out of context approach to seed word selection leads to a conflation of possible literal and metaphorical readings (which we know abound in the domain of expressions relating to the senses). Moreover, the resulted seed word lists do not distinguish related yet clearly distinct senses of the words (i.e, \emph{look} as in "look nice" vs. "look at the painting"). However, given the purpose of this research, this does not constitute a problem since we expect our unsupervised, distributional data-driven approach to capture such nuances and group them in corresponding clusters.

\begin{table}
\centering
\begin{tabular}{ll}
\hline 
\multicolumn{1}{l}{\textbf{Sensory}} & \multicolumn{1}{c}{\textbf{Seed words}} \\ 
\multicolumn{1}{l}{\textbf{Modality}} & \multicolumn{1}{c}{\textbf{}} \\ \hline \hline
\bf{Sight} & see, look, visual, glance, stare, gaze, view, observe, notice, watch \\ \hline
\bf{Hearing} & hear, listen, sound, loud, quiet, soft, audible, audio, voice, silence\\ \hline
\bf{Touch} & touch, feel, sense, sensation, rub, perceive, grasp, press, gentle, light\\ \hline
\bf{Taste} & taste, flavor, savor, savour, palate, bite, mouthful, morsel, eat, teeth \\ \hline
\bf{Smell} & smell, scent, odor, odour, perfume, fragrance, essence, inhale, aroma, olfaction \\ \hline
\end{tabular}
\caption{\label{table-seeds} Samples of 10 seed words per sensory modality considered in this study.}
\end{table}

\subsection{Identifying Sensory Descriptors}

In this subsection we show how we identified and extracted the descriptor words associated with our sensory words of interest (i.e., our expanded seed lists). Although there have been a few recent studies that investigated the semantic and perceptual space of sensory words, to our understanding, these studies have targeted words of taste and/or smell individually in natural language data \cite{Horberg-etal2020,Iatropoulos-etal2018}, and did not analyze their combinations.

In our present work, we go beyond such studies.
Specifically, from the perception corpus, we extracted a set of descriptors identifying content words - i.e., nouns, verbs, adjectives, and adverbs -- for each sense by identifying words that occurred sufficiently many times within a window centered on each seed of each modality list. The context window was cut short if a sentence boundary was encountered - meaning, we considered all words between punctuation marks (periods, commas, colons, semicolons, quotation marks, exclamation points, and question marks) and window boundary. We tested four different window sizes (+/-~4, +/-~10, +/-~15, and +/-~25) and for each size, we experimented with several \textit{cutoffs}, which indicate the number of context windows in which a word must appear in order to be considered a descriptor. This was varied primarily to reduce the number of descriptors identified and ensure that the later computations were computationally feasible in a reasonable amount of time. Increasing the cutoff simply prunes many of the words that appear in some, but not many, context windows. Table \ref{tab:variables} reports the cutoffs tested based on window size. Notice that larger context windows tend to be tested with larger cutoffs. This was intentional as increasing the size of the context window greatly increases the number of candidates for descriptors. The final window size was selected based on the performance of our model and is explained further in Section \ref{sec:model}.

\begin{table}[h]
    \centering
    \begin{tabular}{cl}
    \hline
         \textbf{Window Size} & \textbf{Tested Cutoff Values} \\
         \hline\hline
         $+/- 4\ \ $ & 30, 100, 150, 200, 500, 1000, 2000, 3000 \\
         $+/- 10$ &  300, 400, 500, 1000, 3000\\
         $+/- 15$ &  400, 600, 2000, 6000\\
         $+/- 25$ &  500, 800, 3000, 8000\\
         \hline
    \end{tabular}
    \caption{Breakdown of cutoff values tested based on window size. A cutoff of 30, for example, indicates that a word must appear in at least 30 context windows to be considered a descriptor.}
    \label{tab:variables}
\end{table}

\begin{figure}[h!t]
  \caption{\label{POS200} The distribution of the top 200 descriptors' parts of speech (POS) per each sensory modality. The labels indicate: n (nouns), v (verbs), r (adverbs), and a (adjectives).}
  \centering
    \includegraphics[width=0.6\textwidth]{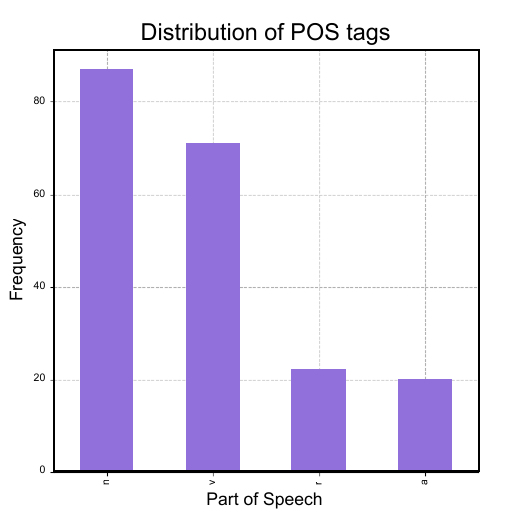}
\end{figure}

Figure \ref{POS200} shows the part of speech distribution of the top 200 most frequent descriptors per each sensory modality. We can see that nouns and verbs predominate in most of the sensory contexts. Note that although we did not classify the original seed words as descriptors, they were selected as such if they passed the descriptor test frequently appearing in context windows of other seeds.

\section{Computational Model}
\label{sec:model}
Taking advantage of the word distributions in the specific sensory contexts around the seed words, we automatically identified sensory descriptors based on the extent to which they express sensory-semantic content. We then derived their semantic organization with distributional-semantic word embeddings.
Specifically, we trained a distributional-semantic word embedding model on all the extracted context windows. In this model, semantic distances between words are represented as vector distances in a multi-dimensional space. Words that occur in similar contexts have similar meanings and therefore also similar vector representations. We experimented with two different embedding models. First, we used a Word2Vec model (CBOW approach) (gensim implementation \cite{vrehruvrek2010software}) with a hidden layer of 200 units, a minimum word frequency of 10, and 20 training iterations. Second, we experimented with a fastText model (CBOW approach) \cite{bojanowski2016enriching} with 100-dimensional word vectors, a minimum word frequency of 5, and 5 epochs. Both models were trained on the set of all context windows with no distinction between/among the senses.
 The distance $D$ between descriptors $i$ and $j$ was calculated as:
\begin{equation}
 D_{ij} = 0.5 * (1 - {p}_{ij}), 
\end{equation}
 where $p$ is the Pearson correlation between word vectors. As such, the distance $D_{ij}$ between descriptors is converted to the 0--1 range, with 0 reflecting semantic identity and 1 indicating semantic opposition. 
Our model allows us to classify words according to the sensory type, and then chart the similarities and differences in the seeds' preference for contextual words. We generated five distance matrices calculated over each individual list of descriptors, one matrix per sense. 

In order to identify clusters of descriptors and to further analyze the primary dimensions along which we can compare the descriptors, we ran the Principal Component Analysis method \cite{PCA} (PCA with 2 principal components) on the five separate distance matrices, each computed with the same embedding model (either Word2Vec or fastText) obtained from all the context windows combined. 
We ran the PCA method for our four possible window sizes to determine which one was most ideal for defining context windows. Table \ref{tab:variance} reports the average variance explained by our models based on window size. This value was obtained via the \texttt{explained\_variance\_ratio\_} field in scikit-learn's PCA model and was averaged over all experiments run with the given window size. The experiments include models with between 2 and 4 principal components and cutoff values indicated by Table \ref{tab:variables}. Additionally, we break down the table based on which embedding model was used to encode the identified descriptors for each model. 

We find that our context window of +/- 4 was able to explain the largest proportion of the variance in the data. We focus on this value as our context window size for the remaining results. Additionally, we chose a threshold of 30 for this window size to reduce the dimensionality of the descriptor distance matrix $D$. We also note that, on average, fastText was able to explain a larger proportion of the variance than Word2Vec, however we focus our analysis on models using Word2Vec which provide more coherent clusters of descriptors upon visual analysis (Section \ref{sec:sensory-blends}). 
\begin{table}[h]
    \centering
    \begin{tabular}{lcc}
    \hline
\multirow{2}{*}{\textbf{Window Size}} & \multicolumn{2}{c}{\textbf{Average Explained Variance Ratio}} \\ \cline{2-3}
& \multicolumn{1}{c}{Word2Vec} & \multicolumn{1}{c}{fastText} \\
\hline\hline
$+/- 4\ \ $     & 0.526  &    0.737            \\
$+/- 10$    &  0.485       &    0.629           \\
$+/- 15$    &  0.486       &     0.654        \\
$+/- 25$    & 0.483         &    \textit{n/a}     \\
\hline
\end{tabular}
    \caption{Average variance explained by PCA models run on descriptors identified using four different window sizes.}
    \label{tab:variance}
\end{table}




\section{Experimental Results and Analysis of the Descriptor Spaces across the Five Sensory Modalities}

In this section, we present experimental results of our further investigation of the differences and similarities of our descriptors' spaces along the basic sensory modalities. 
Such distinctions are extremely important when considering interactions between linguistic, conceptual, and perceptual systems. This is particularly relevant since a common assumption of empirical cognitive studies of perceptual words has been that a concept can be experienced through just one perceptual modality. More recently, researchers in cognitive science have empirically derived measures that allow to classify words as unimodal, bimodal, or multimodal, while separately considering the representational strength on each sensory modality \cite{Lynott2020}.

\subsection{Average Pairwise Distance between Sense Pairs}

\begin{figure}
  \caption{\label{AvgDistPairs} The average pairwise distance between sense pairs in top 200 descriptors. Magenta shows descriptor pairs that belong to the same sense, while gray indicates descriptor pairs of different senses.}
  \centering
    \includegraphics[width=0.7\textwidth]{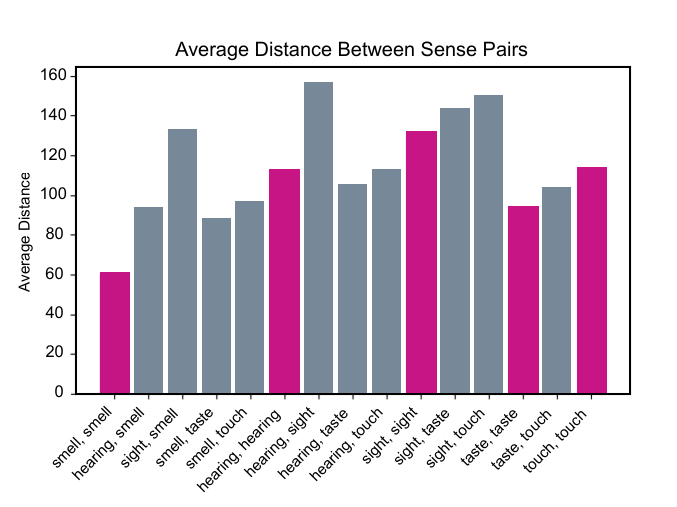}
\end{figure}

We wanted to see how close various descriptors belonging to the same sense but also to different senses were in this space. For this we calculated the pairwise Euclidean distance for every word in the top 200 most frequent words and then averaged the distances for each combination of two senses. Figure~\ref{AvgDistPairs} shows pairs of descriptors belonging to the same sense (in magenta) as well as pairs of descriptors from different sense combinations (in gray). The results indicate that, besides descriptors of smell, as well as those of taste, which tend to cluster together, we also see descriptors of smell--taste as well as smell--touch that share similar contexts. This is not surprising given that the senses of smell and taste are directly related, both using the connected types of receptors.
Sight--hearing, sight--touch, sight--taste did not seem to be so closely related in this space, as given by our top 200 descriptors.

\subsection{Focus points: Sensory Descriptors within a 30-radius Area}

\begin{figure}
  \caption{\label{SensePairs-Radius} The pairs of senses for all top 200 sensory descriptors within a radius of 30 in the generated sensory space. Red shows descriptor pairs that belong to the same sense, while teal indicates descriptor pairs of different senses.}
  \centering
    \includegraphics[width=0.7\textwidth]{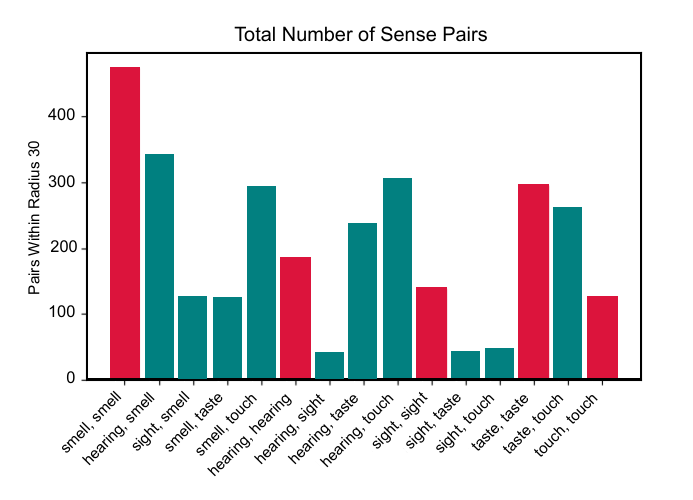}
\end{figure}

We were also interested in analyzing pairs of senses for all data points (i.e., sensory descriptors) within a distance radius of 30.\footnote{We chose the value 30 for the radius based on the results of our previous experiments with average pairwise distance between sense pairs (Figure~\ref{AvgDistPairs}). However, should we had more time for testing, we would have determined the value empirically.} in the generated sensory space. We define here as \emph{focus point} each of the 200 most frequent descriptors. With each descriptor as a focus point, we wanted to see which and how many of other descriptors occurred within its 30-radius area. Figure~\ref{SensePairs-Radius} shows that smell--smell descriptors tend to cluster together more often, followed by hearing--smell and hearing--touch. Sense combinations like hearing--sight, sight--taste, and sight--touch did not seem to cluster together. These findings support the results of the experiments in the previous subsection, Subsection 5.1. Indeed, hearing--sight, sight--touch, and sight--taste do not tend to show as closely related in this space.

\subsection{Sensory Blends: Descriptors Across Sensory Modalities}
\label{sec:sensory-blends}
\begin{figure}
  \caption{\label{SenseOvelap-Multi} The set of 14 descriptors appearing as associated with at least two senses. The set resulted from the sense overlap of the top 200 descriptors.}
  \centering
    \includegraphics[width=0.8\textwidth]{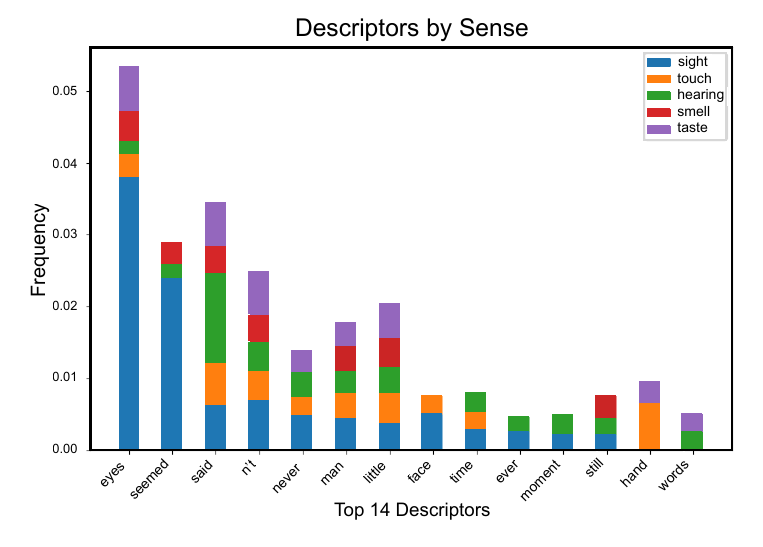}
\end{figure}

We also wanted to see how many words have been seen across different sensory spaces of the five basic senses. For this, we calculated the sense overlap of the top 200 descriptors, considering only descriptors appearing as associated with at least two senses - which gave us a set of 14 such descriptors (i.e., same descriptor appearing with at least two colors - see Figure~\ref{SenseOvelap-Multi}). The frequency with which a descriptor appears with a certain sense was normalized based on how many total context windows there were for a given sense. Figure~\ref{SenseOvelap-Multi} shows that the word \emph{eyes} is seen in association with all the senses, but most often seen as a descriptor of sight. Negations (\emph{n't, never})\footnote{The negation clitic marker \emph{n't} (as in \emph{don't}) is the standard marker representation for contractions used in the Penn Treebank style of tokenization \cite{marcus-etal-1994-penn}.} are seen across the entire sensory spectrum, which was expected. In our top 200 descriptors, \emph{face} has been seen only with sight (e.g., "eyes gazing on a familiar face") and touch (e.g., "winds striking against her face"), while \emph{hand} shows with touch (e.g., "happy touch of her hand") and taste (e.g., "biting deep into the flesh of the hand"). The descriptor \emph{words} is particularly interesting here since it is captured as occurring both with literal (Hearing: "hear her words") as well as metaphoric (Taste: "taste of her bitter, loud words") meanings.

\begin{figure}[ht]
  \caption{\label{PCA-w2v} The PCA score plot of two dimensions PC1 and PC2 scores of the word embedding model of our sensory descriptors. The color codes are as follows: Yellow (sight), Light blue (touch), Red (hearing), Purple (taste), and Dark blue (smell).}
  \centering
    \includegraphics[width=1\textwidth]{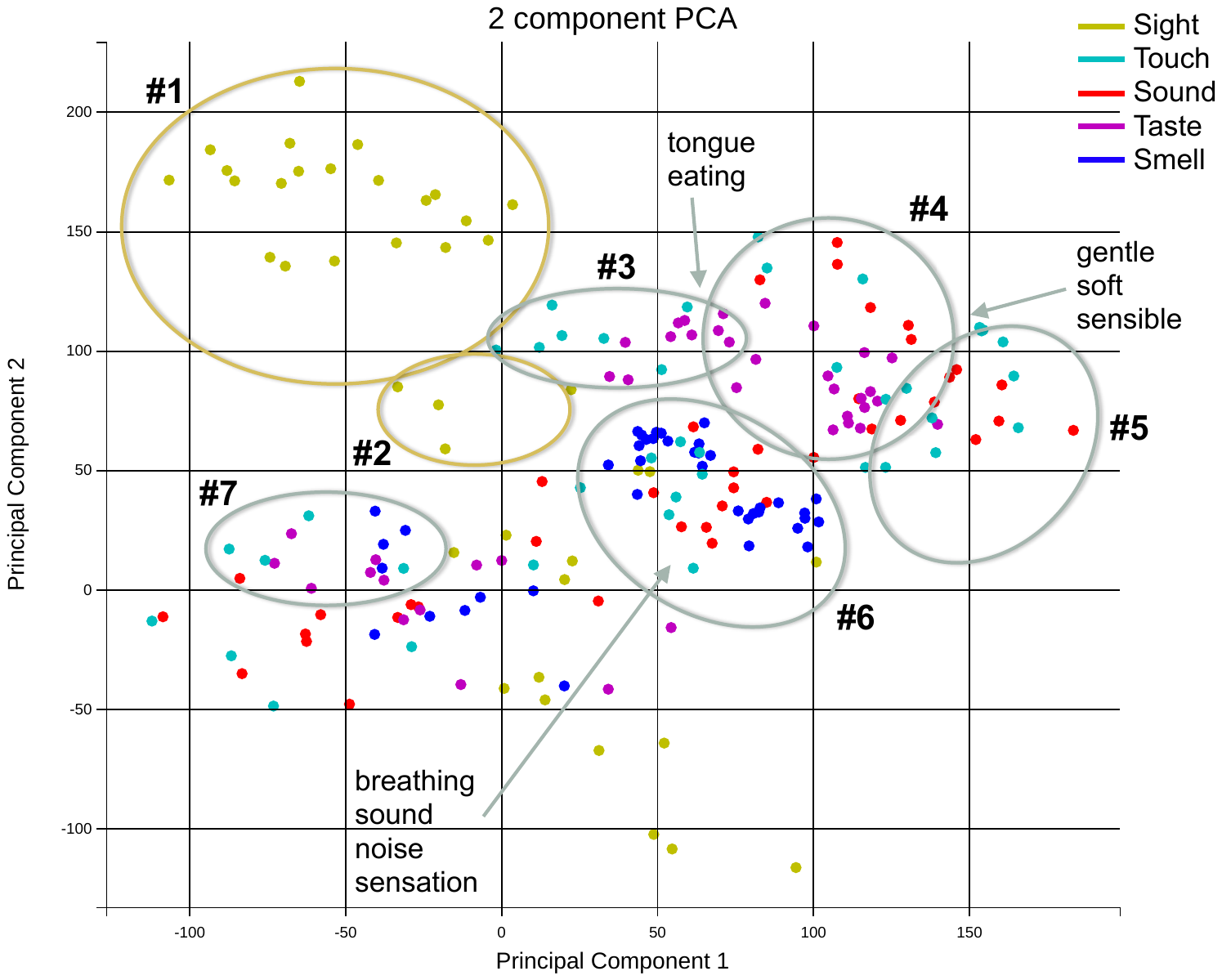}
\end{figure}

\begin{flushleft}
\bf{Sensory blending with PCA}
\end{flushleft}
The two-component PCA results uncover a set of very interesting and informative phenomena. The primary dimension PC1, explains 34.5\% of the variance in the data, whereas the secondary dimension, PC2, 19.65\%. 
As shown in Figure~\ref{PCA-w2v}, descriptors referring to sight (e.g., cluster\#1: \emph{see, notice, appear, observe, look}) tend to have low scores on PC1 of the word embedding PCA model, and high scores on PC2. Both dimensions of PCA analysis seem to be related to sensory associations of various types of descriptors across the modalities.
Specifically, as Figure~\ref{PCA-w2v} shows, while some senses like sight tend to form their own clusters, most senses blend together in a variety of semantic patterns. The sense of sight shows two clearly defined clusters with the bigger cluster (\#1) grouping verbs of sight (e.g., \emph{to see, notice, appear, know, look, watch}), including more specific ways of seeing (e.g., \emph{to stare, glance, gaze}). The sight cluster\#2 captures nouns of sight (e.g., \emph{look, appearance, view}).

However, more interesting are the clusters that form a sensory blending. Consider for instance, cluster\#3 whose members are all body parts (with the exception of the right most one, the verb \emph{eating}). Notice in particular the light blue descriptors on the left of the cluster which refer to body parts of touch (i.e., \emph{finger, fingers} as in "fingers across forehead") as well as adjectives indicating the way fingers move over a surface (i.e., \emph{light, softly}). Toward the center and the right side of the cluster we see semantic shifts toward body parts associated more with taste (purple color - i.e., \emph{skin, tongue, teeth, mouth}), and at the right border, the verb \emph{eating}.

Such purple words of taste associated with ways of eating (in purple in cluster\#3) provide a rather smooth transition into the next cluster, cluster\#4. According to the colors associated with the descriptors, this grouping shows a blend of taste (e.g., \emph{taste, teeth, drink, eating, drinking, swallowing, biting}), touch (as a way of eating: \emph{lightly, gently, feeling}), and hearing (e.g., \emph{quietly, softly, loudly, silently, silence, shattering}) -- i.e., "shattering of teeth in the shadowy silence".
At the bottom right border of cluster\#4 we see red taste words like \emph{gentle, soft, sensible} which provide a nice transition into cluster\#5 that combines touch and sound like \emph{sound, sounded, heard} - i.e., "soft, sweet sound".

One clear tight dark blue cloud of descriptors can be easily spotted in the middle of the diagram. 
This cluster\#6 starts on the upper left side with a few words indicating sense receptors (e.g., \emph{eyes, nostrils}) moving into the top tight dark blue cloud of semantically similar smell words (\emph{fumes, odors, incense, smell, scent, bouquet}) surrounded by light blue words of touch (\emph{touch, feeling, sense, pressure, sensation}). The middle of the cluster is partitioned obliquely by a sequence of red descriptors of sound (\emph{sounds, voices, noise, volume, hearing, breathing, silence}) which transition into the second tight dark blue cloud of more specific smell words (e.g., \emph{breathing, smell, sniffing, wafted}) on the bottom-right side of the cluster. We notice here words like \emph{breathing} showing with two colors - both as a hearing as well as a smell descriptor. The two yellow sight words in this cluster are \emph{watch} and \emph{glance}, while the isolated bottom one is an outlier, \emph{Mr.} Examples of expressions captured by this cluster are "watched the eyes and breathing", "should hear the breathing",  "no further sound broke the stillness", "the strains of martial music faint from distance", "faded sounds", "leaving behind sounds that cried for help". 

In language, we often name words of a particular sense borrowing words from other sensory experiences -- thus, creating potentially new analogies or intimate connections of sensory experience. Take, for instance, the phrase "bitter smell" -- "the words we apply to smells represent either the objects that produce them metonymically, as in \emph{rose}, or their qualities through analogues derived from other modalities, as in \emph{bitter}" \cite{Marks1996}\footnote{Emphasis was the authors' addition.}.

A rather isolated sensory blend cluster is \#7, on the left lower side of the diagram. This combines touch (\emph{face, hand, hands}), taste (\emph{smile, hand, water, words}), and smell (\emph{flower, smoke}). The other bottom left data points on the diagram are negations, as well as outliers like \emph{time, moment, man} which do not belong to any cluster.
Examples of such sensory blending are "can smell and see the smoke in her mouth", "breath his native air", "night air may breath upon us", "hand quickly to his mouth", "biting your hand", "hand over mouth", "words fall from his lips", "words on her lips", "taste of her bitter, loud words".

\section{Conclusion}
Daily, we integrate sight, hearing, smell, taste, and touch as well as our inner sensations and map them to past sensory experiences. The way we interpret this information -- our perceptions -- is what leads to our experiences of the world.

In this paper, we have shown empirical results on the semantic organization of English sensory descriptors based on a corpus of over 8,000 fiction books from Project Gutenberg. 
We have applied a distributional-semantic word embeddings method that automatically identifies sensory descriptors in natural texts as how likely they are to occur in contexts within and across the five basic senses. Our method also provides the semantic organization of the identified descriptors given their distribution in those contexts. Of course, as with any word embedding approach, our results are predicated on a number of parameters of the model, including the seed words we started with. The extent to which the model is sensitive to these seed words needs to be further explored - i.e., investigate and test various seed lists at different levels of abstraction and compare with other dictionaries or ontologies, and evaluate this aspect in more detail. It would also be interesting to see whether the findings of the analysis are confirmed by more complex models like transformers (i.e., BERT \cite{devlin2018pretraining}, etc.). Overall, a big advantage of our approach is that it can be easily applied to other languages and to different texts, since it is fully unsupervised. 

The results and insights obtained from our experiments are based on fiction novels written or translated in English. It would be interesting to see how do our extracted sets of sensory descriptors compare with existing normative sensory words lists in the literature (e.g., lists obtained by Lievers and colleagues \cite{Lievers-etal2013} and Lynott and colleagues \cite{Lynott2020}).
Another possible and interesting test would be to see if similar findings can be obtained for other languages. Many humanities researchers might also be interested in adding and evaluating a diachronic dimension of this research. Indeed, such an analysis would potentially show interesting usage of sensory language and how it might differ from one century to another.

This study is part of a larger project which focuses on writers’ narrative strategies prompting readers to use and combine senses thus creating sensory images that bring characters and scenes to life. This line of research can increase our understanding of blending senses and may be used to inform creative writing, as well as to better our understanding of the ways the senses work to engage the reader of literary texts. Such insight may also help in a better understanding of the sensory system, overall. We believe the extensive analysis on sensory blending presented in this paper is a good starting point to uncover the strength of the sensory language in fiction and in creative language in general. 
In future work, we intend to expand the analysis along the following directions.

\begin{flushleft}
\bf{ 1) Multiperspective Fiction}
\end{flushleft}
Multiperspectivity is a characteristic of narration, where the event or the story is told from multiple viewpoints. Most frequently the term is applied to fiction which employs multiple narrators, thus establishing a deeper sense of interconnectivity. One of the writer's challenges here is to use language effectively to move to one point of view to the next. Usually, this is done through a sensory shift where a sense like sight, sound, or smell is perceived by more than one narrator. 
The writer invites the reader to enter the characters' minds inferring and tracing characters’ mental and emotional states in order to make sense of their behavior through a blend of sight, sound, touch, taste, and smell. Shared sensations thus can lead narrative perspective from one character to the next revealing who these people are, with their hopes and wishes. Despite the significant interest in multiperspectivity, there are still few narratological works devoted to this research landscape \cite{Hartner2014}.

\begin{flushleft}
\bf{2) Cross-sensory spaces of literal and metaphorical context}
\end{flushleft}
Metaphors are entrenched in language, suggesting color and imagery, helping us to understand complex concepts we may not be familiar with, to connect with each other, and even to shape our thought processes. Metaphor and perceptual experience are tied together in narrative literature \cite{Lakoff-Johnson1981}. Anthropologist Brenda Beck defined metaphors as "bridges", arguing that "If forced to delimit the concept of metaphor I would insist on the experiential, body-linked, physical core of metaphorical reasoning abilities" \cite{Beck1987}.

Sensory descriptions, as we have seen in this paper as well, can often occur in both figurative and literal contexts, yet with different behavior. The different ways that sensory descriptors are employed by the writer as well as processed by the reader can offer insights into cultural and social understandings of the senses and the ways in which the human sensory apparatus works.

Many perceptually based metaphorical expressions (e.g., "a faded sound", "a sweet smell") seem to be connected to the structure of perceptual experiences and sensory system. In his work on perceptual metaphors, Marks \cite{Marks1996} argued that "Even if some perceptual metaphors end up being mediated linguistically, their origins appear to be wholly in perception itself, starting within perceptual processes before being overlaid and dominated by linguistic ones". 
Although there has been significant progress in the metaphor arena, less attention has been given to how fiction characters make sense of the world through different senses, and to how the visual, hearing, touch, taste, and smell metaphors differ in the kind of knowledge they encode \cite{Otis2015}.
This is another great topic for future research.

\bibliographystyle{elsarticle-num-names} 
\bibliography{cas-refs}





\end{document}